\pdfoutput=1

\documentclass[11pt]{article}

\usepackage[preprint]{acl}

\usepackage{times}
\usepackage{latexsym}
\usepackage{makecell}

\usepackage[T1]{fontenc}

\usepackage[utf8]{inputenc}

\usepackage{microtype}

\usepackage{inconsolata}
\usepackage{multirow}
\usepackage{booktabs}
\usepackage{caption}

\usepackage{graphicx}

\title{Advacheck at GenAI Detection Task 1: AI Detection Powered by Domain-Aware Multi-Tasking}

\author{
 \textbf{German Gritsai\textsuperscript{1, 2}},
 \textbf{Anastasia Voznyuk\textsuperscript{1}},
 \textbf{Ildar Khabutdinov\textsuperscript{1, 2}},
 \textbf{Andrey Grabovoy\textsuperscript{1}}
\\
 \textsuperscript{1}Advacheck OÜ, Estonia\\
 \textsuperscript{2}Université Grenoble Alpes, France\\
 \texttt{\{gritsai, voznyuk, khabutdinov, grabovoy\}@advacheck.com}
}

\begin{document}
\maketitle
\begin{abstract}
The paper describes a system designed by Advacheck team to recognise machine-generated and human-written texts in the monolingual subtask of GenAI Detection Task 1 competition. Our developed system is a multi-task architecture with shared Transformer Encoder between several classification heads. One head is responsible for binary classification between human-written and machine-generated texts, while the other heads are auxiliary multiclass classifiers for texts of different domains from particular datasets. As multiclass heads were trained to distinguish the domains presented in the data, they provide a better understanding of the samples. This approach led us to achieve the first place in the official ranking with 83.07\% macro $F_1$-score on the test set and bypass the baseline by 10\%. We further study obtained system through ablation, error and representation analyses, finding that multi-task learning outperforms single-task mode and simultaneous tasks form a cluster structure in embeddings space. We release our code and model\footnote{\url{https://github.com/Advacheck-OU/ai-detector-coling2025}}.
\end{abstract}

\section{Introduction}
With the continuous improvement of Large Language Models (LLMs), the task of detection machine-generated texts demands more and more attention from the community. The potential cases of misuse include malicious usage by students~\cite{zeng2023automatic, koike2023outfox} and scientists~\cite{ma2023ai, aij23}. Furthermore, this is often the cause of plagiarism~\cite{plag} and spam~\cite{spam}. The mentioned things are encouraging researchers to improve methods for detecting artificial text simultaneously with enhancing generation methods.

The task of detection of the machine-generated texts is usually formulated as a binary text classification task~\cite{jawahar-etal-2020-automatic}. The most common solutions are to fine-tune the Transformer-based model~\cite{iber} or to use zero-shot approaches with intrinsic statistics of the text~\cite{DetectGPT, zeroshot}. While these methods perform well on in-domain tasks~\cite{TURINGBENCH}, they are not robust to change of the domain, generator model, or language of the texts~\cite{wang2023m4, Tulchinskii_phd, kuznetsov2024robustaigeneratedtextdetection}. Meanwhile, for the detection of AI-content in the wild such a change is, on the contrary, a more realistic setup~\cite{raid}. Beyond this, the data presented for the detection task may be of poor quality, which also complicates the challenge of the detection task~\cite{dataquality}. Therefore, the goal is to obtain a model that is robust to the presence of data of poor quality and with a lot of noise, and, in addition, to make this model capable to adapt to new domains. 

\begin{figure}[t]
  \centering
  \includegraphics[width=0.4\textwidth]{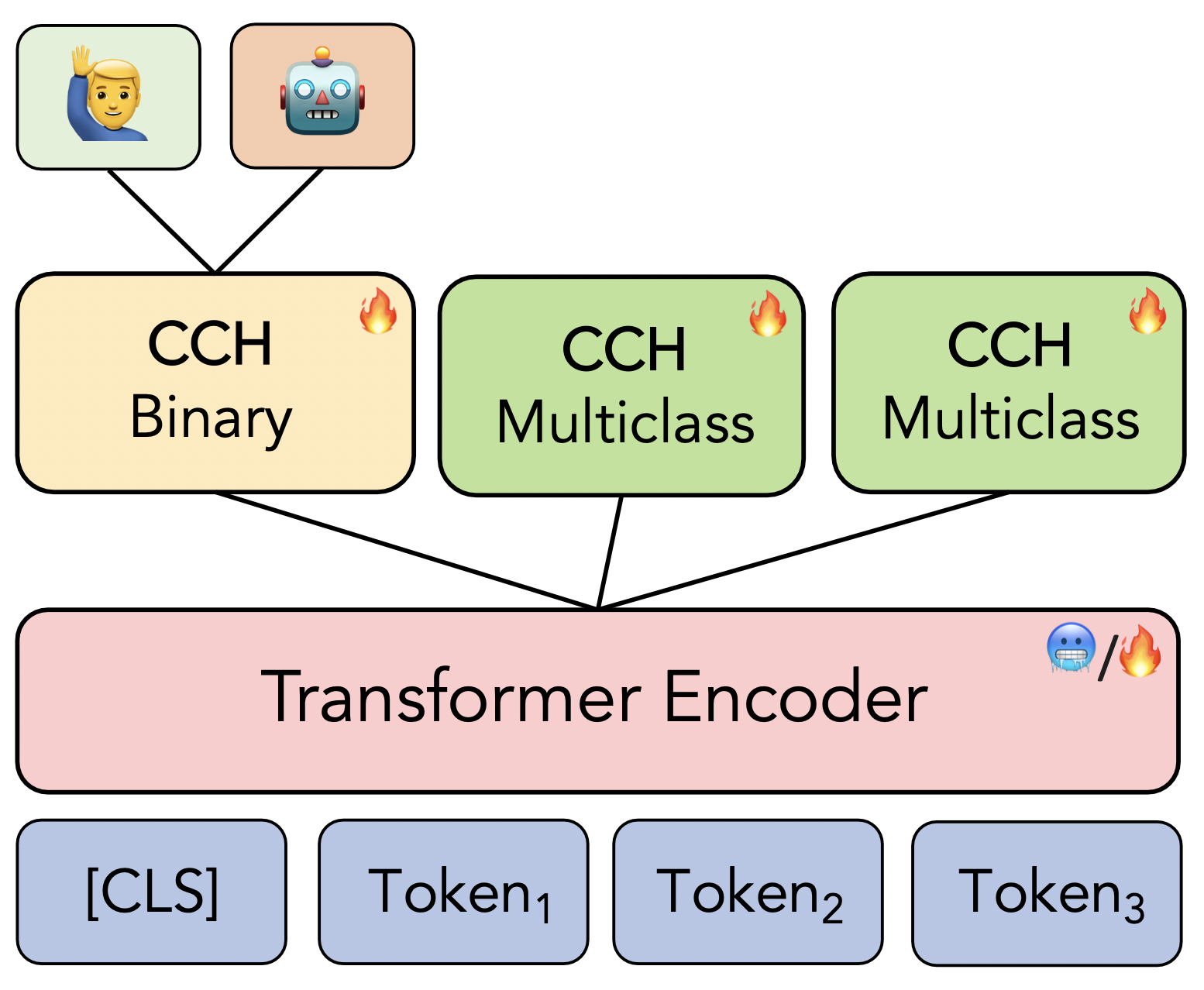}
  \caption{Overview of the proposed multi-task architecture. Modules marked only with \raisebox{-0.25em}{\includegraphics[scale=0.015]{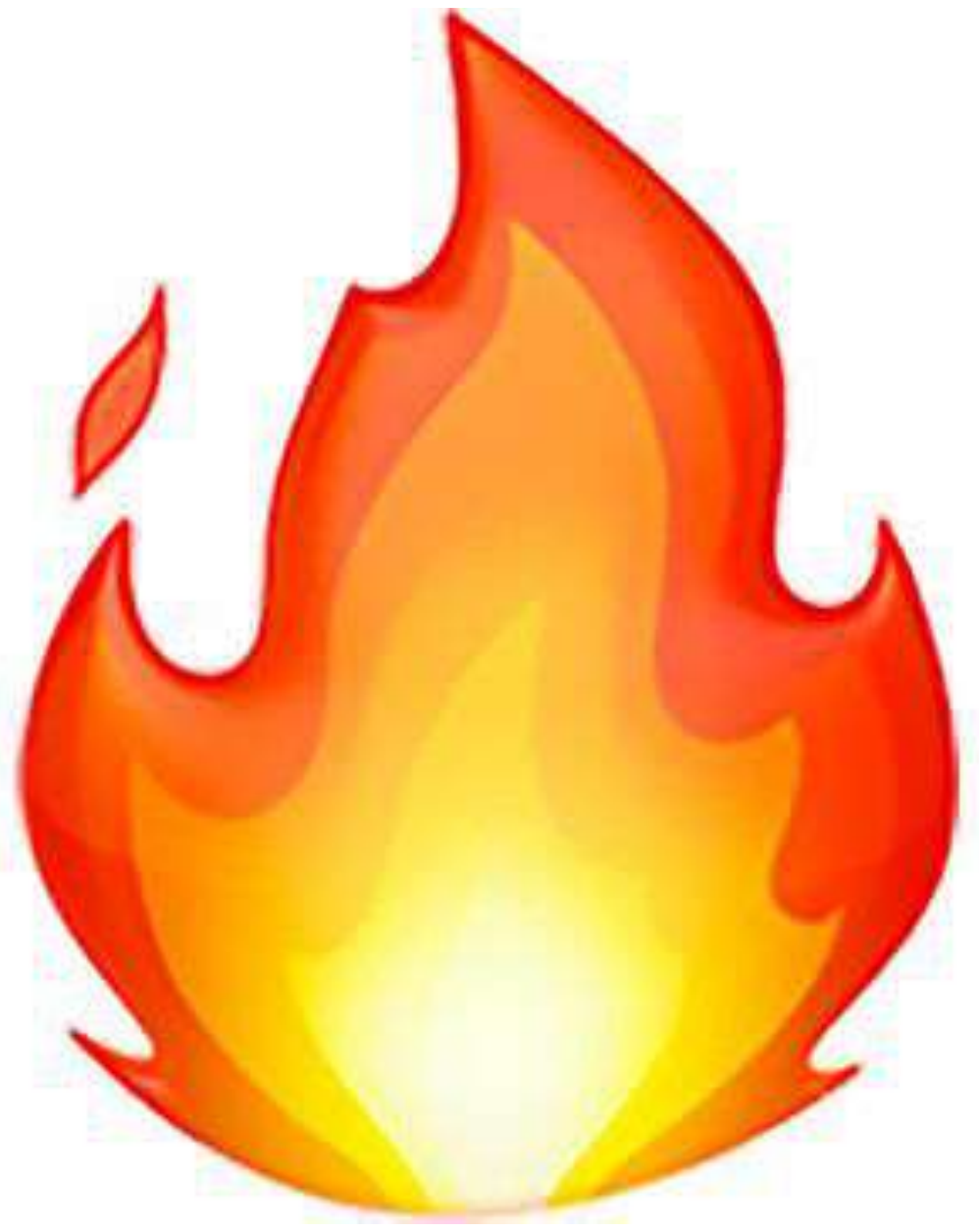}} are trainable at all stages. The weights of Transformer Encoder are frozen \raisebox{-0.25em}{\includegraphics[scale=0.015]{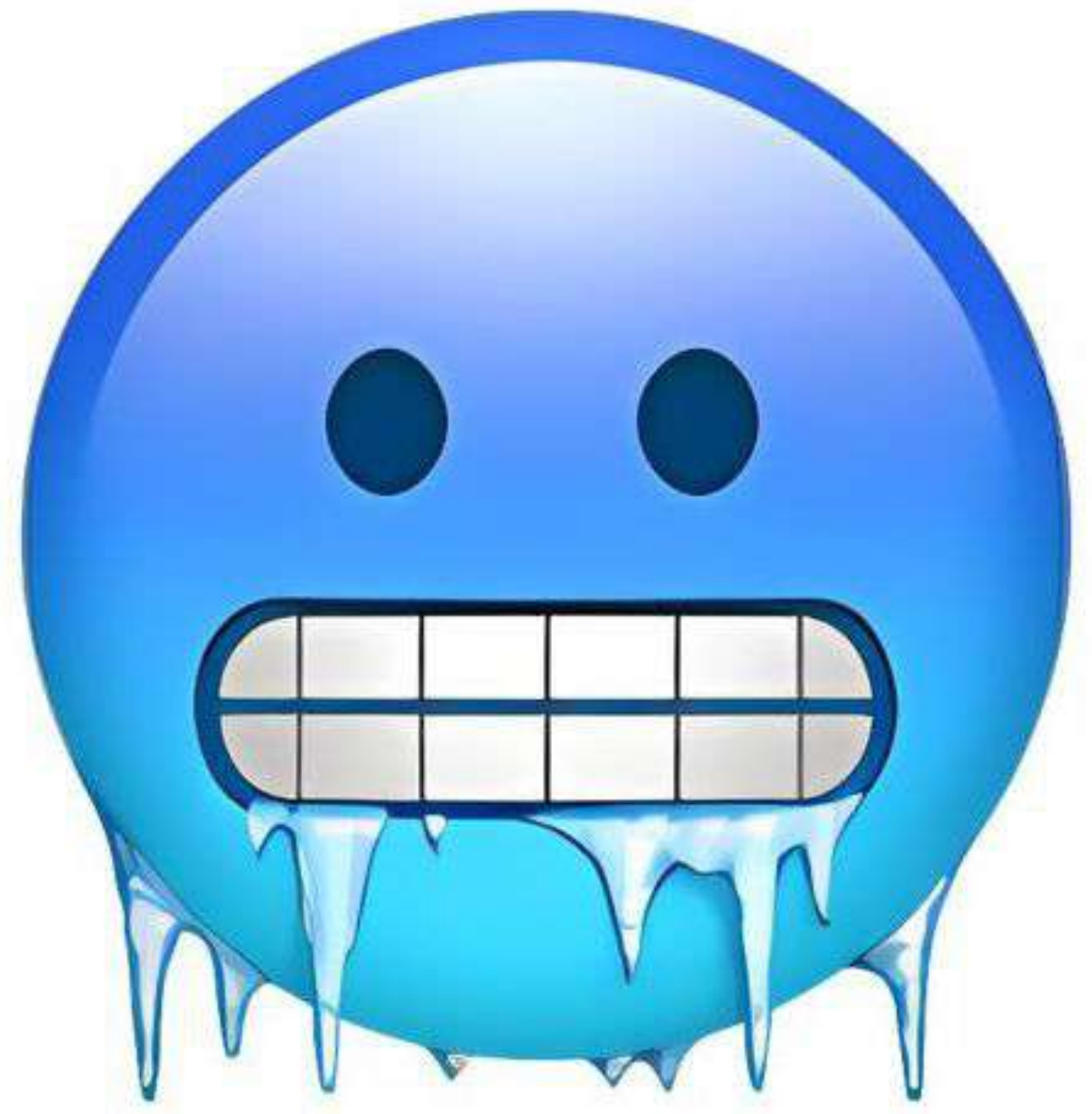}} at the first stage of training and trainable \raisebox{-0.25em}{\includegraphics[scale=0.015]{images/fire.pdf}} at the second one. The Custom Classification Head (CCH) described in Appendix \ref{sec:appendix_a} is used for predictions.}
  \label{fig:pipeline}
\end{figure}

The high coherence and quality of writing achieved by modern LLMs makes it difficult to find a specific sample-counted qualitative feature by which it would be possible to create a hyperplane in the space of texts and separate generated from human-generated ones. One of the possible refinements of the representations from the single-task learning architecture is the multi-task learning (MTL)~\cite{mtl, needmore}. It was also noted that systems with MTL architecture had achieved high results in the previous competitions~\cite{mtlsemeval}, therefore we decided to utilize this approach in our work.
In this paper we discuss our solution as the Advacheck team at GenAI Content Detection Task 1: English and Multilingual Machine-generated Text Detection: AI vs. Human~\cite{wang2025genai}. Our method shows that with additional internal data analysis and embedding alignment using MTL, it is still possible to achieve high performance in detecting fragments in cross-domain and cross-generator setups on texts from the advanced LLMs. As we forced model to focus on various domains, it allowed us to form a cluster domain-wise structure for the text representations in the vector space. In our research, we show that (1) multi-task learning outperforms the default single-task, (2) cluster structure is formed at the shared encoder (3) compare different configurations of the system and (4) analyse the errors of the system.

\section{Task Definition}

The monolingual subtask of \textit{Task 1: Binary Multilingual Machine-Generated Text Detection} focuses on identifying whether the English text was entirely authored by a human or generated by a language model. The competition is the continuation and improvement of the \textit{SemEval Shared Task 8 (subtask A)}~\cite{semeval2024task8} and combines refreshed training and testing samples from different domains and novel LLMs. The statistics of the dataset are summarised in Appendix \ref{sec:appendix_c}. The official evaluation metric for the monolingual subtask is Macro $F_1$-score and the additional metric is Micro $F_1$-score.

\section{System Overview}
\textbf{Why multi-task learning?} In the current task formulation we have more than 600k texts in the training set with dozens of domains and generation models. Such an amount is very noisy for a binary classification task, because it can be challenging for a model to distinguish between relevant and irrelevant features. Multi-task learning may help the model focus on those features that actually matter as other tasks will update representation of samples with inner information. Our aim is to obtain fine-grained representations of the data that ideally ignores data-dependent noise and generalises well. Since different tasks involve distinct noises, a model trained on multiple tasks simultaneously is able to learn a more general representation. Furthermore, it reduces the risk of overfitting.

\textbf{Model.} We propose a MTL architecture with hard parameter sharing (HPS), it is depicted in Figure \ref{fig:pipeline}. In HPS, a common Transformer-based encoder is used for multiple tasks. After several variations of set of parallel heads, we focused on three custom classification heads (CCH) for simultaneous training:
\begin{itemize}
    \item Binary CCH head for solving the initial monolingual subtask [2 classes]
    \item Multiclass CCH to define a sub-source within the HC3~\cite{Su2023HC3PA} source [5 classes]
    \item Multiclass CCH for sub-source detection within the M4GT~\cite{wang2023m4} source [6 classes]
\end{itemize}

The model was trained in two phases: fine-tuning chosen classifiers with frozen \raisebox{-0.25em}{\includegraphics[scale=0.015]{images/snow.pdf}} shared encoder weights and fine-tuning the complete model with all weights unfrozen \raisebox{-0.25em}{\includegraphics[scale=0.015]{images/fire.pdf}}. At the inference stage, only binary CCH predictions used for final classification.

\section{Experiments}

\begin{figure*}[t]
  \includegraphics[width=1.03\linewidth]{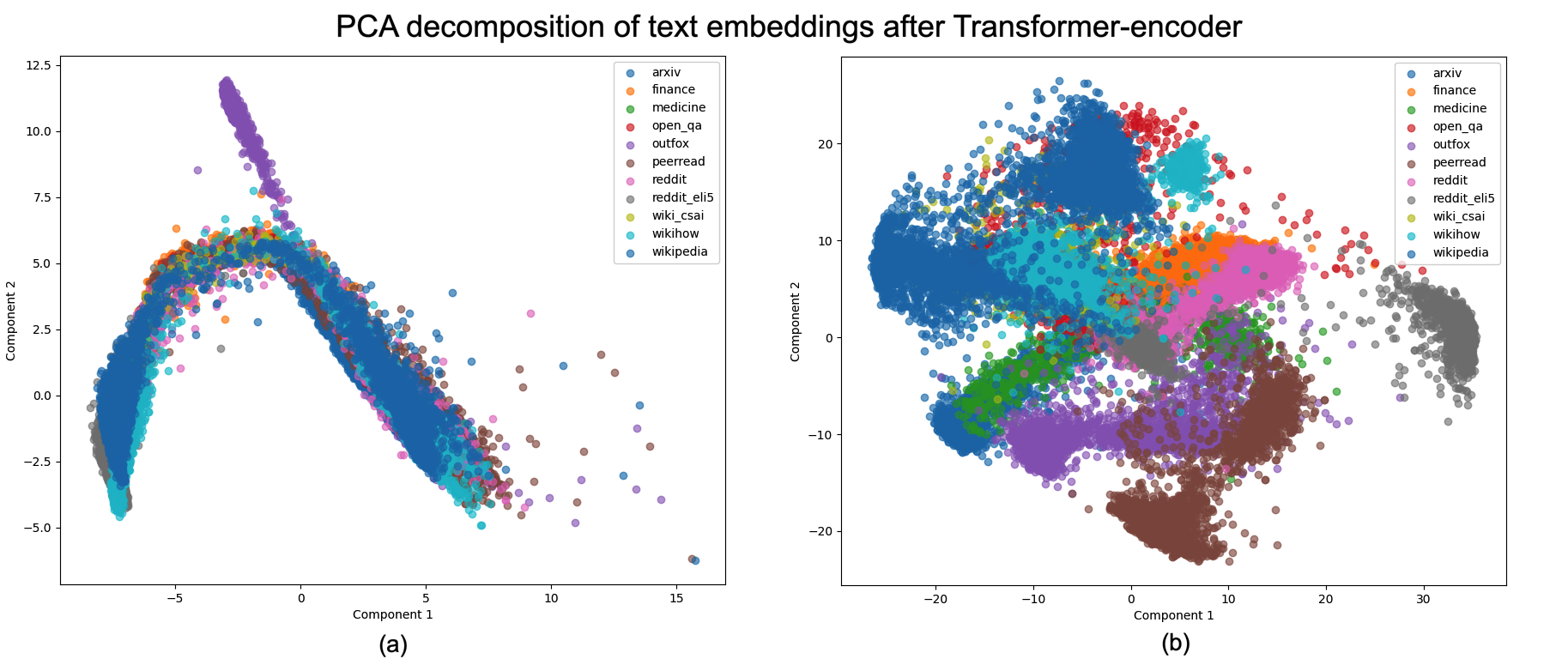} \hfill
  \vspace{-15pt}
  \caption {Two principal component decomposition of PCA for texts from the development subsample. In (a) the vector space structure for the \texttt{deberta-v3-base} fine-tuned in single-task mode is shown, while figure (b) shows the same model but fine-tuned in MTL mode with two additional custom classification heads.}
  \label{fig:pca}
\end{figure*}

We focused on the monolingual subtask, carrying out comparisons among models and ablations of the best system. For these we employed the original training and development splits provided by the organizers. Our objective here was to reveal the quality improvement in multi-task training compared to single-task training.

\begin{table}
  \centering
  \begin{tabular}{lcc}
    \toprule
    \textbf{Model} & \textbf{Development} & \textbf{Test}\\
    \hline
    TF-IDF with LogReg & 63.53 & 60.93 \\
    DeBERTaV3 base  & 82.56 & 78.52\\\hline
    MTL: 1 stage & 80.51 & 78.67 \\
    MTL: 2 stage & 87.33 & 81.55 \\
    MTL: 2 stage + threshold & \textbf{87.96} & \textbf{83.07} \\\bottomrule
  \end{tabular}
  \caption{Results of model comparison on the test and development set. The highlighted metric is macro $F_1$-score (\%).}
  \label{tab:comp}
\end{table}

\subsection{Model Comparison}
\textbf{Baselines.} As models for comparison we chose Logistic Regression classifier \cite{logreg} with TF-IDF features on word n-grams, and DeBERTa-v3 fine-tuned in two-stage mode described earlier, but in single-task setting. In the MTL approach, we compared checkpoints from different stages, and also explored the effect of adding thresholds on the output of the final classifier.
We chose DeBERTa-v3 base for the baseline and the backbone in our system, as it is currently state-of-the-art model for supervised fine-tuning  for binary classification~\cite{Macko_2023}.

The results are presented in Table \ref{tab:comp}. It can be seen that there is a weak correlation between the gap within the predictions on the dev and test subsamples. For example, the presence of a threshold after the final layer affected the dev result only slightly, but at the same time allowed us to achieve a winning result on the test set. The hyperparameters of the final model are given in Appendix \ref{sec:appendix_b}.

\begin{table}[h!]
  \centering
  \begin{tabular}{rlc}
    \toprule
    \textbf{Rank} & \textbf{System} & \textbf{$F_1$-score (\%)} \\
    \hline
    \textbf{1} & \textbf{Advacheck (germgr)}  & \textbf{83.07} \\
    2 & tmatchitan &  83.01 \\
    3 & karla  & 82.80 \\
    \hline
    15 & \textit{baseline} & 73.42 \\
    \hline
    36 & nitstejasrikar & 44.89 \\
    \bottomrule
  \end{tabular}
  \caption{Final results on the official ranking. Bold
denotes our team’s placement.}
  \label{tab:results}
\end{table}

\section{Results}

Table \ref{tab:results} reports the leaderboard results on the test set, where our system, Advacheck, achieves a macro $F_1$-score of 83.07\%, outperforming approaches of the other participants and ranking first. Our solution surpassed the claimed baseline by 10\%.

\begin{figure}[t]
  \centering
  \includegraphics[width=\linewidth]{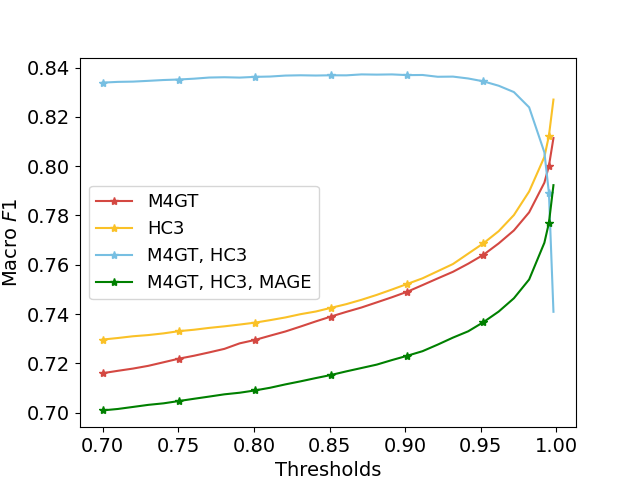}
  \caption{Macro $F_1$-score on the test set of different configuration of the systems depending on the threshold.}
  \label{fig:thresholds}
\end{figure}

\section{Analysis}

\subsection{Embeddings after MTL} We made a comparison of text embeddings after fine-tuning stages. Samples from development part of data were forwarded to the Transformer-encoder and \texttt{[CLS]} vectors were extracted as outputs. We visualised these vectors using PCA in Figure \ref{fig:pca}. We observe that the alignment of the representations, which was mentioned above, introduces a cluster domain-wise structure. Although the clusters are not perfectly separable, a meaningful difference between the standard BERT-like model and MTL fine-tuning pipelines can be seen. Additional decompositions are presented in Appendix \ref{sec:appendix_e}.

\subsection{Ablations}

To provide further analysis on the multi-task setup we experimented with configurations of our systems, changing the number of multiclass CCH. The system on the leaderboard has 2 Multiclass CCH, and we ran ablation experiments with 1 CCH and 3 CCH. The results are in the Table~\ref{tab:ablations}. Setups with 1 and 3 CCH showed better performance on the development set, but marginally dropped in performance on the test set compared to setup with 2 CCH. What is also interesting is that the results obtained on HC3-trained CCH are similar to the results obtained in M4GT-trained CCH, although M4GT has 10 times more training data than HC3. Additionaly, we experimented with the threshold values on all our configurations. The figures are shown in the Figure~\ref{fig:thresholds} and reaffirm the choice of the final system and threshold for it.

\begin{table}[t]
\resizebox{\columnwidth}{!}{%
\begin{tabular}{lcc}
\toprule
\textbf{Task Head} & \textbf{Development} & \textbf{Test} \\
                  \hline
HC3               &  92.27 &  82.70 \\
M4GT              &  91.70 & 81.07 \\
\hline
MTL (HC3 + M4GT)  &  87.96  & 83.07  \\
\hline
HC3 + M4GT + MAGE & 91.43  & 79.23  \\
\bottomrule
\end{tabular}%
}
\caption{Comparison of different configurations of heads and tasks trained simultaneously in MTL architecture. The highlighted metric is macro $F_1$-score (\%).}
\label{tab:ablations}
\end{table}

\subsection{Error Analysis}

\textbf{Answers on different datasets.} The two datasets with the highest percentage of incorrect predictions are Mixset~\cite{zhang2024llmasacoauthor} and CUDRT~\cite{tao2024cudrtbenchmarkingdetectionmodels}, while the texts in the other datasets are detected with very high precision. We attribute this to the additional manipulations with these texts, such as rewriting, ``humanizing'' and other editing, done by the authors of the initial datasets. Other texts in the test set and in train set, on the other hand, are the raw output of generation models, without post-processing them, or otherwise not stated. See more details in Appendix~\ref{sec:appendix_d}.

\begin{figure}[t]
  \centering
  \includegraphics[width=\linewidth]{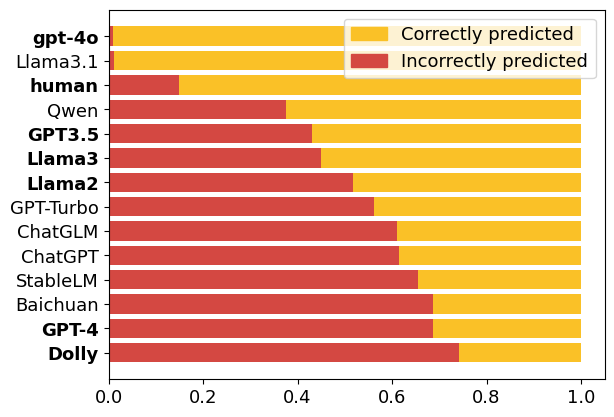}
  \caption{Proportion of predictions for different generators from test set. Labels in \textbf{bold} are generators texts from which are present in the train set.}
  \label{fig:proportion_generators}
\end{figure}

\textbf{Answers on different generators.} The proportions of correct and incorrect predictions on test set with respect to different generators are shown in Figure~\ref{fig:proportion_generators}. The majority of texts in the test set are either human-written or generated with \texttt{gpt-4o} and our system predicted labels for them very accurately, therefore compensating the poorer performance on other generators. Still, our system is not robust enough yet to the change of generators, as on texts from some of unseen detectors and we dedicate our future works to it.

\section{Conclusion}

In this paper we described the system by the Advacheck team in the monolingual subtask at GenAI Detection Task 1 competition. We proposed solution with multi-task learning architecture that consists of shared Transformer Encoder and composition of one binary and two multiclass Custom Classification Heads. Our system obtained the best results in the official ranking bypassing the baseline by 10\%. Adding tasks for training in parallel reveal the formation of a cluster structure in the space of embeddings, helping to achieve high results despite the presence of a large amount of noisy data. Also, it has been demonstrated that training a similar model but in single-task mode loses to the proposed approach, and configurations with one or three multiclass heads also perform worse than our final system.

\bibliography{main}

\appendix

\section{Custom Classification Head}
\label{sec:appendix_a}

\begin{figure}[t]
  \centering
  \includegraphics[width=0.36\textwidth]{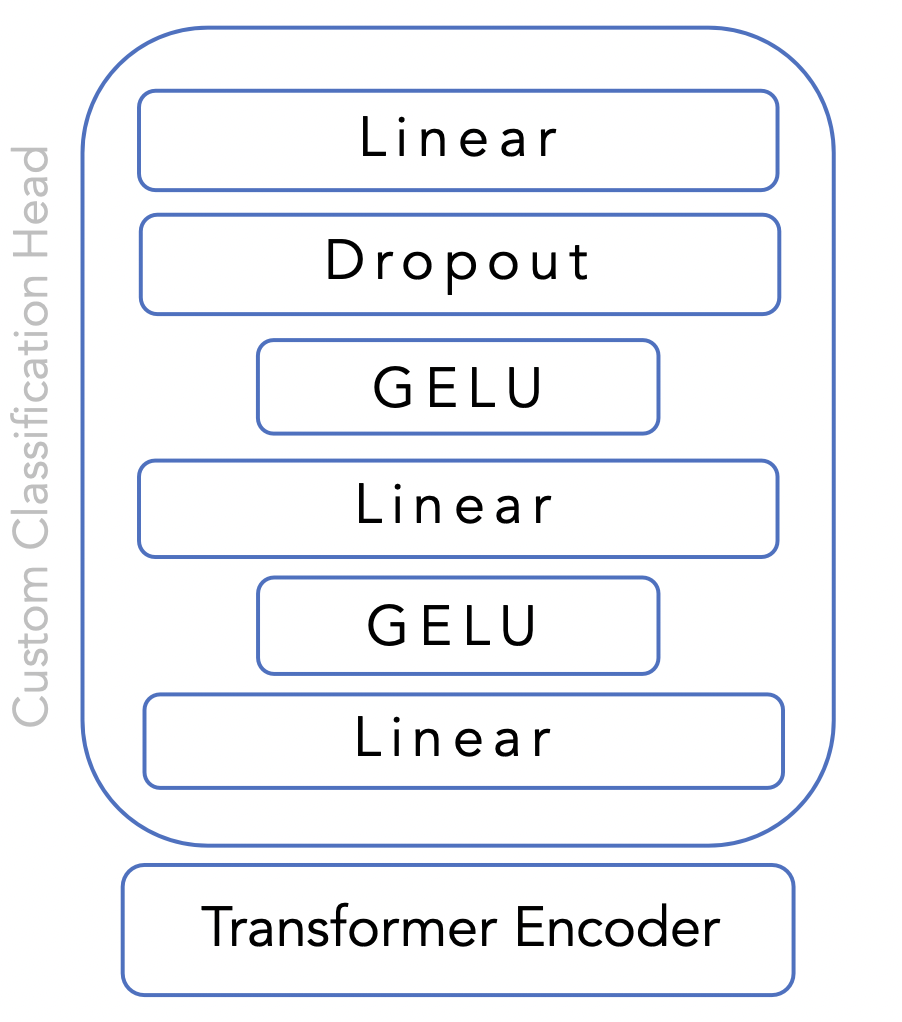}
  \caption{The architecture of the custom classification head used in the described approach. Value of \texttt{dropout} is equal to 0.5.}
  \label{fig:head}
\end{figure}

In our approach, we replaced the default one-layer linear classifier with a more extended version by adding multiple layers, the final structure of Custom Classification Head (CCH) is shown in Figure \ref{fig:head}. We chose GELU \cite{gelu} as the activation feature and added dropout. In earlier experiments, when compared with the base head, this adaptation gives a higher quality therefore we used it in all subsequent experiments.

\section{Final System Hyperparameters}
\label{sec:appendix_b}
\label{appendix:hyperparameters}
\begin{table}[h!]
\centering
\begin{tabular}{lcc}
\toprule
\textbf{Hyperparameters} & \textbf{1st stage} & \textbf{2nd stage}\\
\hline
Epochs & 1 & 1 \\
Learning rate (LR) & 3e-4 & 3e-6 \\
Warmup steps & 50 & 75 \\
Weight decay & 0.01 & 0.01 \\
Batch size & 32 & 16 \\
Classifier threshold & - & 0.92 \\
\bottomrule
\end{tabular}
\caption{Hyperparameters for fine-tuning MTL architecture. We trained for 1 epoch in both stages with possibility of early exit.}
\label{tab:hyperparameters}
\end{table}

Our final system is MTL architecture with shared \texttt{deberta-v3-base} encoder and 3 CCH. The training of main system was conducted on NVIDIA GeForce RTX 3090 and the training of other configurations on NVIDIA A100. See hyperparameters in Table~\ref{tab:hyperparameters}.

\section{Provided Data}
\label{sec:appendix_c}

The organizers of the competition provided data for the train and development stages of the evolving solutions. This is the continuation and improvement of the SemEval Shared Task 8 samples. New domains and generation models were added to the data; details of train and dev sets are shown in Table \ref{fig:data}. In addition, a separate development dataset was available on the CodaBench\footnote{\url{https://www.codabench.org/competitions/3734/}} platform where the competition was held; its statistics are shown in Table \ref{fig:data_dev}.

\begin{table}[h!]
\centering
\renewcommand{\arraystretch}{1.3} 
\begin{tabular}{lrr}
\toprule
\textbf{Source} & \multicolumn{2}{c}{\textbf{Development Set}} \\
                & \textbf{Human} & \textbf{Machine} \\
\midrule
RAID             & 13371 & 0 \\
\hline
LLM-DetectAIve          & 0 & 19186 \\
\midrule
\textbf{Total}  &  & 32557 \\
\bottomrule
\end{tabular}
\caption{Statistics on development data from CodaBench platform for monolingual subtask of the GenAI Detection Task 1.}
\label{fig:data_dev}
\end{table}

\begin{table}[h!]
\centering
\renewcommand{\arraystretch}{1.3} 
\begin{tabular}{lrr}
\toprule
\textbf{Source} & \multicolumn{2}{c}{\textbf{Test Set}} \\
                & \textbf{Human} & \textbf{Machine} \\
\midrule
CUDRT            & 12287          & 10691            \\
\hline
IELTS Duck       & 9747           & 12418            \\
\hline
PeerSum          & 5080           & 6995             \\
\hline
LLM-DetectAIve   & 1635           & 900              \\
\hline
Mixset           & 0              & 1086             \\
\hline
NLPeer           & 5326           & 5376   \\  
\bottomrule
\end{tabular}%
\caption{Statistics on test data for monolingual subtask of the GenAI Detection Task 1.}
\label{tab:test_set}
\end{table}

\begin{table*}[t]
\centering

\renewcommand{\arraystretch}{1.3} 
\begin{tabular}{lp{7.0cm}rrrr}
\toprule
\textbf{Source} & \textbf{Sub-sources} & \multicolumn{2}{c}{\textbf{Training Set}} & \multicolumn{2}{c}{\textbf{Dev Set}} \\
                &                     & \textbf{Human} & \textbf{Machine} & \textbf{Human} & \textbf{Machine} \\
\midrule
HC3             & \makecell[l]{Finance, Medicine, OpenQA, Reddit\_ELI5, \\ Wiki\_CSAI}   & 39140 & 17671 & 16501 & 7917 \\
\hline
M4GT            & \makecell[l]{Arxiv, Outfox, PeerRead, Reddit,\\  WikiHow, Wikipedia}   & 86682 & 180381 & 36420 & 74167 \\
\hline
MAGE            & \makecell[l]{CMV, CNN, DialogSum, ELI5, HellaSwag,\\ IMDB, PubMed, Roct, SciGen, SQUAD, \\ TLDR, WP, XSum, Yelp }& 103100 & 183793 & 45407 & 81462 \\
\midrule
\multicolumn{2}{r}{\textbf{Total}} & 228922 & 381845 & 98328 & 163430 \\
\bottomrule
\end{tabular}
\caption{Statistics on training and development data from monolingual subtask of the GenAI Detection Task 1.}
\label{fig:data}
\end{table*}

\section{Percentage of Failures}
\label{sec:appendix_d}
\begin{figure}[h!]
  \centering
  \includegraphics[width=0.9\linewidth]{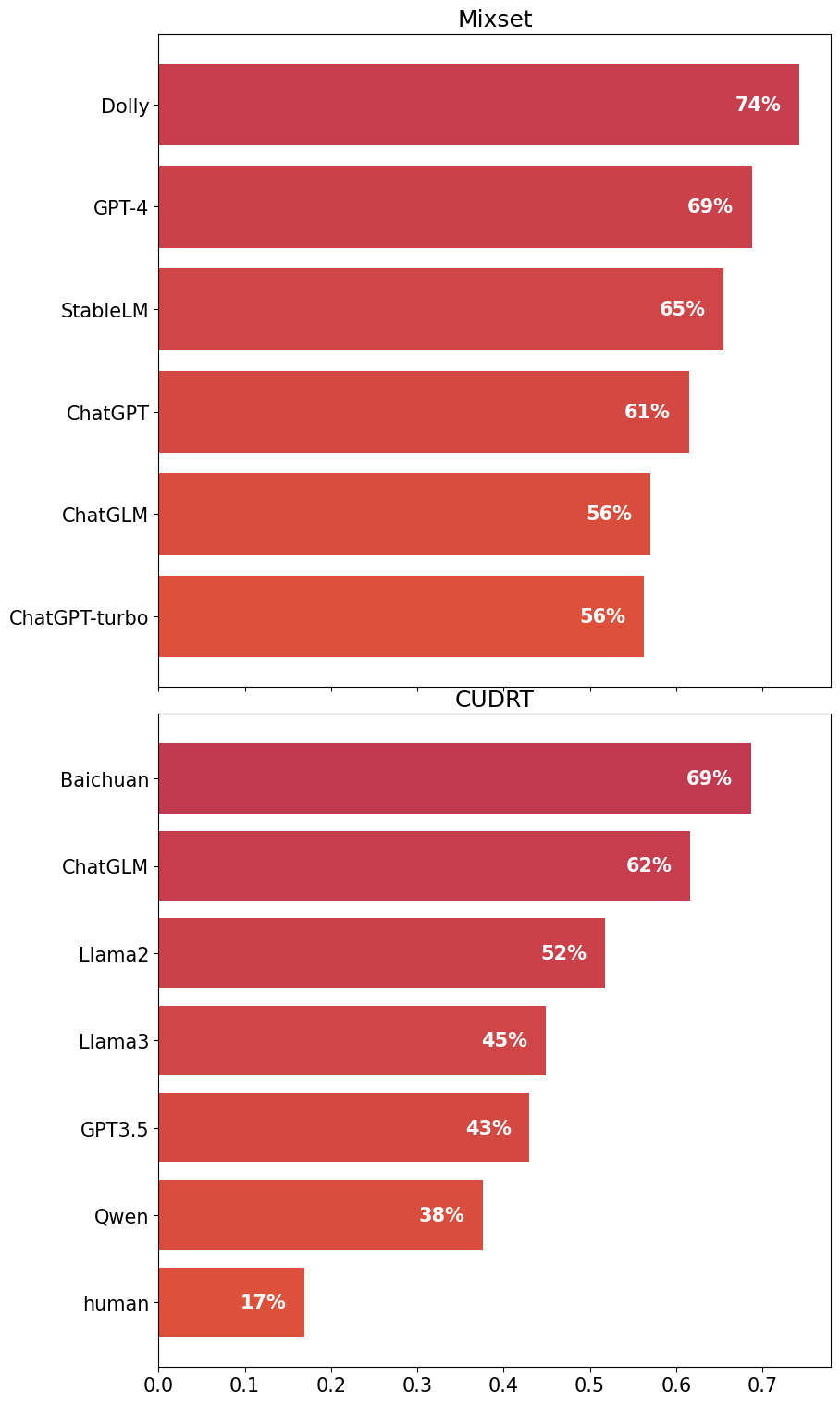}
  \caption{The percentage of falsely labelled texts from two subdatasets in test set.}
  \label{fig:percentage}
\end{figure}

See Figure~\ref{fig:percentage} for detailed proportions of incorrectly predicted texts from CUDRT and Mixset.

\section{Decomposition Study}
\label{sec:appendix_e}
\begin{figure*}[t!]
  \includegraphics[width=1.03\linewidth]{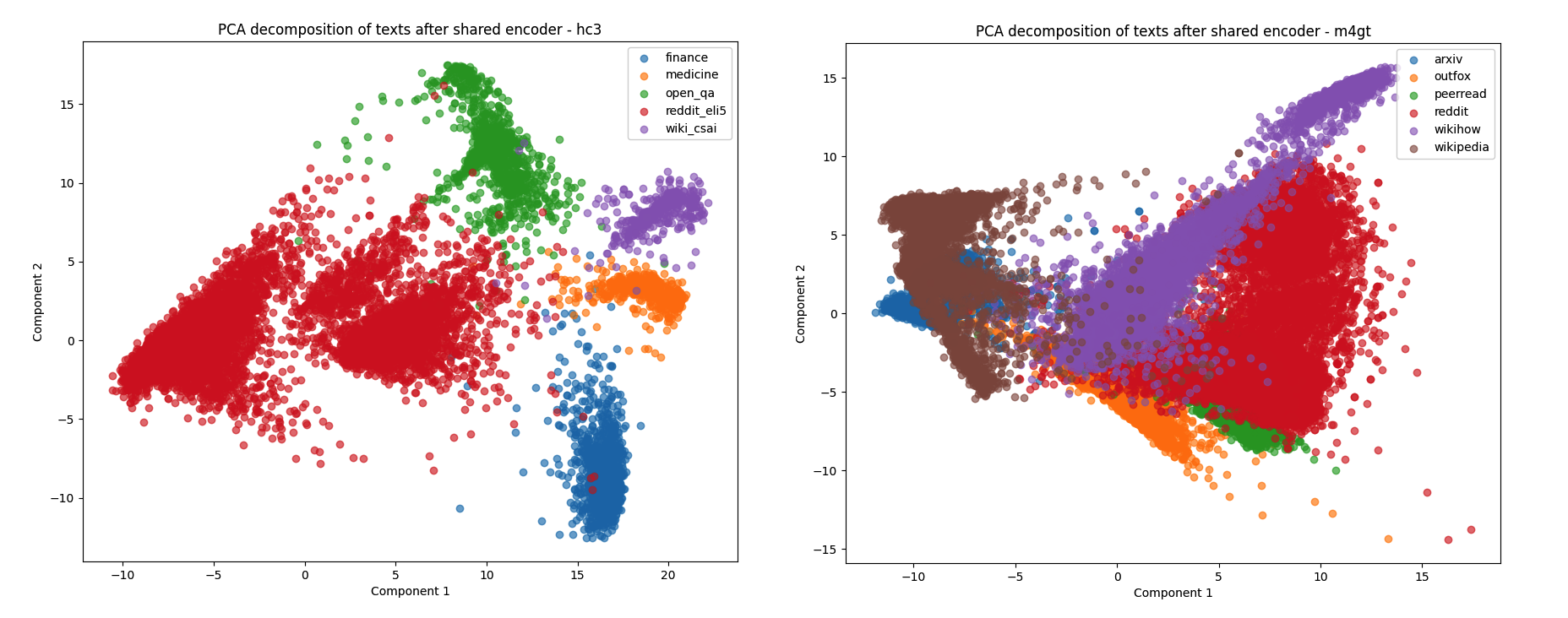} \hfill
  \vspace{-15pt}
  \caption {Two principal component decomposition of PCA for texts from development sub-sample. We decomposed here by PCA the logits of the texts after the corresponding multiclass classifiers, namely HC3 and M4GT.}
  \label{fig:pca_add}
\end{figure*}

In addition to the PCA decomposition of text embeddings after passes on our system, we mapped the logit decomposition of two multiclass heads -- CCH on HC3 and CCH on M4GT. We sampled the texts from the dev set, passed them through the encoder and the corresponding classifiers, and then decomposed the logits. From Figure \ref{fig:pca_add}, we can observe that the data after the classifiers passes remain in the expected cluster structure intended by shared encoder. On the dataset HC3 this can be seen more clearly. With these plots we can also understand how well the multiclass classification heads were trained directly, as they were not used for the inference.

\end{document}